\documentclass[letterpaper]{article} 
\usepackage{aaai25}  
\usepackage{booktabs}
\usepackage{bm}
\usepackage{amssymb}
\usepackage{multirow} 
\usepackage{footnote}
\usepackage{xcolor}
\usepackage{arydshln}

\usepackage{xcolor}

\usepackage{times}  
\usepackage{helvet}  
\usepackage{courier}  
\usepackage[hyphens]{url}  
\usepackage{graphicx} 
\usepackage{natbib}  
\usepackage{caption} 
\usepackage{algorithm}
\usepackage{algorithmic}

\usepackage{xcolor}

\usepackage{newfloat}
\usepackage{listings}
\DeclareCaptionStyle{ruled}{labelfont=normalfont,labelsep=colon,strut=off} 
\floatstyle{ruled}
\newfloat{listing}{tb}{lst}{}
\floatname{listing}{Listing}
\title{Stable-Hair: Real-World Hair Transfer via Diffusion Model}
\author {
    Yuxuan Zhang\textsuperscript{\rm 1},
    Qing Zhang\textsuperscript{\rm 6}, \\
    Yiren Song\textsuperscript{\rm 4}, 
    Jichao Zhang\textsuperscript{\rm 3},
    Hao Tang\textsuperscript{\rm 5},
    Jiaming Liu\textsuperscript{\rm 2 \textdagger}
}
\affiliations {
    \textsuperscript{\rm 1}Shanghai Jiao Tong University,
    \textsuperscript{\rm 2}Tiamat AI, 
    \textsuperscript{\rm 3}Ocean University of China,
    \textsuperscript{\rm 4}National University of Singapore, \\
    \textsuperscript{\rm 5}Peking University
    \textsuperscript{\rm 6}Shenyang Institute of Automation Chinese Academy of Sciences, \\
}

\begin{document}

\twocolumn[{
\renewcommand\twocolumn[1][]{#1}%
\maketitle
\begin{center}
    \centering
    \includegraphics[width=0.98\textwidth]{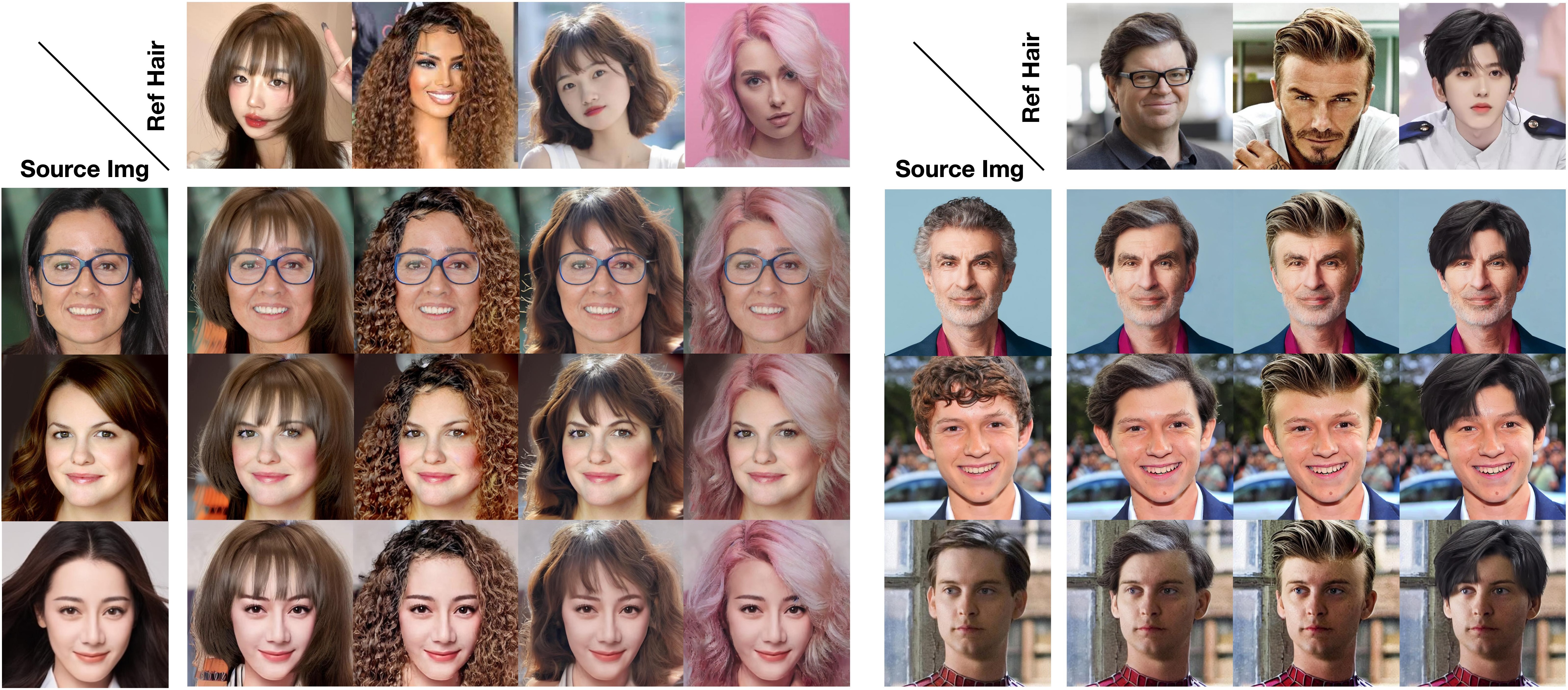}
    \vspace{-0.2cm}
    \captionof{figure}{Stable-Hair is the first diffusion-based method for hairstyle transfer, capable of handling an extensive range of real-world hairstyles with exceptional robustness. Unlike previous methods, which often struggle with complex or intricate styles, Stable-Hair achieves remarkably detailed and high-fidelity transfers while preserving the original identity content.}
    \label{fig:teaser}
\end{center}
}]

\footnotetext{\textdagger: Corresponding author.}

\begin{abstract}
Current hair transfer methods struggle to handle diverse and intricate hairstyles, limiting their applicability in real-world scenarios. In this paper, we propose a novel diffusion-based hair transfer framework, named \textit{Stable-Hair}, which robustly transfers a wide range of real-world hairstyles to user-provided faces for virtual hair try-on. To achieve this goal, our Stable-Hair framework is designed as a two-stage pipeline. In the first stage, we train a Bald Converter alongside stable diffusion to remove hair from the user-provided face images, resulting in bald images. In the second stage, we specifically designed a Hair Extractor and a Latent IdentityNet to transfer the target hairstyle with highly detailed and high-fidelity to the bald image. The Hair Extractor is trained to encode reference images with the desired hairstyles, while the Latent IdentityNet ensures consistency in identity and background. To minimize color deviations between source images and transfer results, we introduce a novel Latent ControlNet architecture, which functions as both the Bald Converter and Latent IdentityNet. After training on our curated triplet dataset, our method accurately transfers highly detailed and high-fidelity hairstyles to the source images. Extensive experiments demonstrate that our approach achieves state-of-the-art performance compared to existing hair transfer methods.
\end{abstract}
\section{Introduction}
Hair transfer is one of the most challenging tasks in the virtual try-on domain. The objective of this task is to transfer hair color, shape, and structure attributes from a reference image to a user-provided source image while preserving the identity and background of the source image. In recent years, advances in Generative Adversarial Networks (GANs)~\cite{tan2020michigan,2018arXivhair-Gans,2022CtrlHair,2022HairNet,2023HairNeRF,chung2022hairfit,zhu2021barbershop,saha2021LOHO,wei2022hairclip,wei2023hairclipv2,nikolaev2024hairfastgan,Khwanmuang2023StyleGANSalon,kim2022styleSYH} have driven significant progress in this field. However, these GAN-based methods often struggle to handle the diverse and complex hairstyles encountered in real-world scenarios, severely limiting their effectiveness in practical applications. Recently, diffusion models have emerged as SOTA methods in the field of image generation. These models not only enable more stable training but also demonstrate impressive results in terms of diversity and fidelity. Consequently, we are intrigued by the question: \textit{``Can we leverage the powerful capabilities of diffusion models to achieve more stable and high-precision hair transfer?"}

Building on the aforementioned considerations, in this paper, we propose a novel hair transfer framework based on diffusion models, named Stable-Hair. Inspired by method~\cite{wei2023hairclipv2} which converts different editing conditions (e.g., text, sketch) into different proxies in the StyleGAN $\mathcal{W}$+
space, we designed a two-stage paradigm to ensure the precision and naturalness of our hair transfer process. In the first stage, we train a Bald Converter to transform the user-provided source image into a bald proxy image. Furthermore, a key advancement in our methodology is the implementation of an automated data generation pipeline, which generates triplets for training. This pipeline utilizes the Large Language Model for generating text prompts, and the Stable Diffusion Inpainting model for creating reference images. This synthetic training data ensures the effective training of our framework, enabling it to robustly handle challenging hairstyles with remarkable fidelity. In the second stage, a Hair Extractor is specially trained to capture and encode hairstyles from reference images with unprecedented levels of detail and texture. A Latent IdentityNet is used to encode the source image, ensuring that the identity and background of the source image remain consistent throughout the transfer process. By integrating these two components with diffusion U-Net, we can effectively inject the features of the reference hair into the source image. This approach not only allows the hair to adapt accurately to the new environment of the source image but also ensures that it is styled appropriately, appearing natural and matching the original appearance of the subject.

To rigorously preserve the non-hair regions and maintain the facial identity of the subject, we have developed an innovative architecture named Latent ControlNet. This structure integrates our Bald Converter and Latent IdentityNet. By transforming the pixel-space transfer process to operate within the latent space of the diffusion model, Latent ControlNet ensures color consistency in the non-hair regions of the source image during the transfer process. 

Through extensive experiments, Stable-Hair has demonstrated its superior performance, significantly surpassing existing state-of-the-art hair transfer methods in terms of fidelity and fine-grained detail. Our approach sets a new standard for hair transfer technology, promising to revolutionize the virtual hair try-on experience with its advanced capabilities and innovative design.

In summary, our contributions are: 
\begin{itemize}
    \item In this paper, we propose the first diffusion-based hair style transfer framework, named \textit{Stable-Hair}. Experiments demonstrate that our method significantly surpasses existing SOTA hair transfer methods in terms of fidelity and fine-grained detail.
    \item We utilize a Hair Extractor to encode reference images and inject detailed hair features. To ensure better source content consistency during the transfer process, we introduce a novel Latent ControlNet architecture. This architecture is utilized as a Bald Converter and a Latent IdentityNet, which maps the hairstyle transfer process from the pixel space to the latent space.
    \item We utilized our designed pipeline to process both video and image data. This pipeline plays a crucial role in the effective training of our framework by generating a high-quality dataset.
    
\end{itemize}
\section{Related Works}

\subsection{Hair Style Transfer}
The development of GAN-based methods \cite{tan2020michigan,2018arXivhair-Gans,2022CtrlHair,2022HairNet,2023HairNeRF,chung2022hairfit,zhu2021barbershop,saha2021LOHO,wei2022hairclip,wei2023hairclipv2,nikolaev2024hairfastgan,Khwanmuang2023StyleGANSalon,kim2022styleSYH,shu2022few} has significantly advanced the field of hairstyle transfer, with most current methods relying on GAN-based approaches.

\begin{figure*}[h]
    \centering
 \includegraphics[width=1.\linewidth]{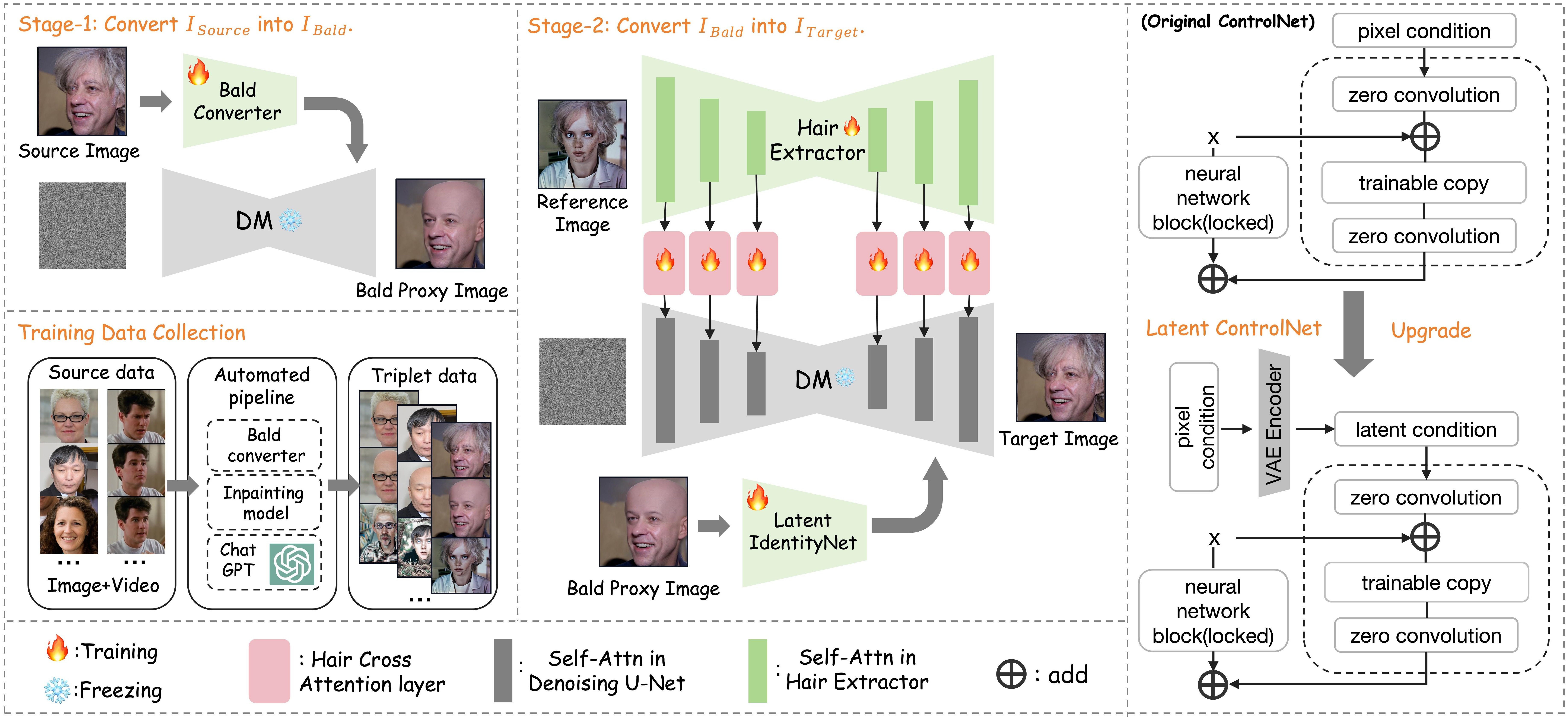}
    \caption{Overall schematics of our method. Our pipeline consists of two stages. First, the user's input source image is transformed into a bald proxy image by utilizing a Bald Converter. In the second stage, we employ the pre-trained SD model along with a Hair Extractor to transfer the reference hair onto the bald proxy image. The Hair Extractor is responsible for capturing the intricate details and features of the reference hair. These features are then injected into the SD model through newly added hair cross-attention layers. After training on the triplet dataset constructed using our specially designed automated data pipeline, our method achieves highly detailed and high-fidelity hair transfers, resulting in natural and visually appealing outcomes.}
    \vspace{-0.3cm}
    \label{fig:method}
\end{figure*}

In hair transfer, MichiGAN \cite{tan2020michigan} decomposes hair into four orthogonal attributes and designs the corresponding modules to represent, process, and convert user input, which is then integrated to realize an end-to-end network. Barbershop \cite{zhu2021barbershop} proposes a novel latent space for image blending that excels at preserving detail and encoding spatial information, extracting hairstyle information from multiple reference images. Furthermore, LOHO \cite{saha2021LOHO} employs an optimization-based approach using GAN inversion to infill missing hair structure details in the latent space during hairstyle transfer, introducing two-stage optimization and gradient orthogonalization to enable disentangled latent space optimization of hair attributes. Hairmapper \cite{Wu_2022_CVPR} trains a network to direct hair removal in StyleGAN's latent space. Although these early methods can transfer simple hairstyles, they do not address the challenge of pose-invariant hairstyle transfer. Therefore, SYH \cite{kim2022styleSYH} proposes a pose-invariant model that handles significant pose differences while maintaining local textures. Hairclip \cite{wei2022hairclip} utilizes CLIP for unified hair editing, with HairCLIPv2 \cite{wei2023hairclipv2} converting hair editing tasks into transfer tasks with varied proxies. StyleGAN-Salon \cite{Khwanmuang2023StyleGANSalon} employs multi-view optimization and guide images to enhance hairstyle transfer accuracy. HairFastGAN \cite{nikolaev2024hairfastgan} addresses the challenge of hairstyle transfers in different poses and proposes a new encoder capable of generating images promptly.

Despite these advancements, GAN-based models have limitations in addressing complex hairstyle transfers in real-world scenarios. To overcome these deficiencies, we propose the first diffusion-based method, which generates high-quality and robust images, thereby setting a new benchmark in the field.

\subsection{Diffusion Models}
Currently, diffusion models have attracted a great deal of attention and have seen significant advancements. As the most prominent generative models today, diffusion models have achieved state-of-the-art results across various image generation tasks, including text-to-image generation~\cite{dalle2, sdxl, Imagen, rombach2022high, IF}, image editing~\cite{attendandexcite, layerdiffusion, selfguidance, masactrl,editevery,dragondiffusion,iedit,sine,tsaban2023ledits,yang2024editworld,li2024zone}, controllable generation~\cite{controlnet,t2i,unicontrol,directed}, personalized image generation~\cite{zhang2024ssr, lora, domainagnostic,taming,TI, hyperdreambooth, DB,zhang2024fast} and so on. In particular, the recent emergence of diffusion-based virtual try-on applications further demonstrates the formidable generative capabilities of diffusion models, enabling commercial grade virtual try-ons~\cite{xu2024ootdiffusion,stablegarment,kim2023stableviton,zeng2023cat} and virtual makeovers~\cite{zhang2024stable}, which were previously unattainable with traditional GAN methods.

Diffusion models excel in generating high-quality images, especially when trained on extensive datasets. In this paper, We use diffusion models to achieve high-fidelity and robust hairstyle transfers, addressing the limitations of GAN-based methods and significantly enhancing the quality and realism of complex hairstyles.

\section{Methodology}
\subsection{Overview}
 As shown in Fig.~\ref{fig:method}, our design divides the hairstyle transfer process into two stages. Firstly, the user's input source image is transformed into a bald proxy image by using the Bald Converter. Secondly, our model is used to transfer the reference hair onto the bald proxy image. This two-stage approach ensures optimal stability in hairstyle transfer and maintains the source image content. 
 
 Specifically, given a reference image containing the desired hairstyle and a source image provided by users, Hair Extractor can stably and precisely encode a variety of reference hairstyles in real-world scenes and extract detailed hair features. These detailed hair features are fed to the diffusion U-Net structure. The Latent IdentityNet modules are used to process the bald proxy image that is converted from the source image by the Bald Converter. 

\subsection{Stage1: Standardizing to Bald State}
\subsubsection{Bald Converter.} In our method, the source image is processed in two steps, the first step is to convert the source image into a bald proxy image by a \textit{Bald Converter}. Specifically, we use the vanilla diffusion model in conjunction with our designed Latent ControlNet structure (introduced in the following section) to transfer the source image into the bald state. 

After training, the Bald Converter achieves complete hair removal without requiring image alignment or cropping. This results in a cleaner, more consistent baseline for subsequent hairstyle transfer, leading to improved visual fidelity and more realistic outcomes. Notably, standardizing source images to a bald state is crucial for enhancing model performance in hairstyle transfer. This process allows the model to concentrate more effectively on facial features and details by removing hair-related variability. Such standardization is instrumental in improving both training stability and transfer accuracy. The significance and impact of this approach are discussed in greater detail in the supplementary materials.

\begin{figure}
    \centering
    \includegraphics[width=1\linewidth]{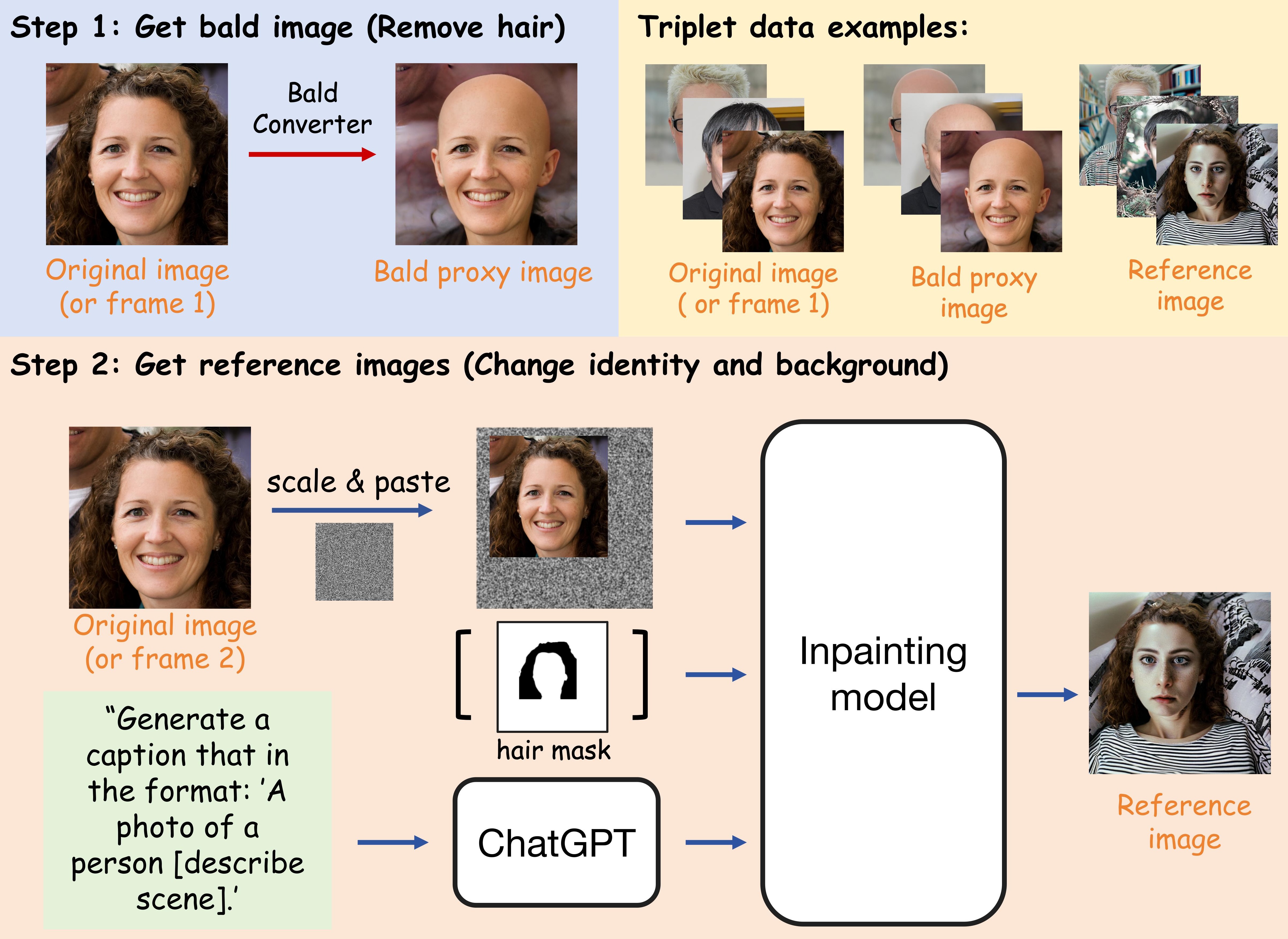}
    \vspace{-0.4cm}
    \caption{\textbf{Synthetic Training Data:} We propose an automated data generation pipeline to generate \{Original image (or frame 1), Reference image, Bald proxy image\} triplets for training. The pipeline uses ChatGPT to generate text prompts, the Stable Diffusion Inpainting model to generate reference images, and our pre-trained Bald converter to convert the original image or one of the frames sampled from videos into the bald proxy image.}
    \vspace{-0.3cm}
    \label{fig:data}
\end{figure}

\subsection{Training Data Collection}
After training our Bald Converter, we can utilize the Bald Converter alongside the Stable Diffusion Inpainting model and ChatGPT to generate the triplet training data necessary to train the second-stage models.

\subsubsection{Automatic Pipeline.} As shown in Fig.~\ref{fig:data}, our pipeline to create the hairstyle pairing dataset involves two main steps. First, we use the bald converter to generate bald images, which serve as proxy images for bald heads during training. Second, based on the hairstyle masks from the original images, we use a Stable Diffusion inpainting model and ChatGPT to edit the non-hairstyle portions of the original dataset, altering identities and backgrounds to create the reference images for training. Ultimately, we obtain a triplet dataset consisting of the original images, the reference images, and the bald proxy images. 

\subsubsection{Video Strategy.}
Relying solely on image data, the reference and source images typically share the same head pose. This limitation prevents the model from effectively handling cases where the head poses in the reference and source images are misaligned. Video data, on the other hand, often includes frames of the same individual (with consistent hairstyle) from different angles. By using video data to generate our dataset—selecting one frame as the source image and another with a different pose as the reference image—we can elegantly address this issue. This approach enhances the model's robustness to pose variations.

In terms of image data, we used every single image as original images to generate reference images and bald proxy images, resulting in a total of 60,000 images. For video data, we sampled two frames from the same video. One frame was used to generate bald proxy images, and another frame was used to generate reference images. Resulting in a total of 90,000 images.

\subsection{Stage2: Hair Transfer and Integration}
\subsubsection{Latent IdentityNet.}
In our method, the second step is to transfer the reference hair onto the bald proxy image. The maintenance of content consistency in the source image is crucial through our two steps. Any deviation from the intended goal, such as alterations in color or identity, would result in a content-inconsistent final image. Therefore, designing a maintenance module becomes a pivotal aspect of our hair transfer framework. 

A baseline approach is using the ControlNet structure as a \textit{Bald Converter} and \textit{Latent IdentityNet} to ensure content consistency. However, our experimental findings indicate that while ControlNet effectively maintains the structural consistency of the source image, it struggles to preserve color consistency. As shown in Fig.~\ref{fig:ablation}, due to accumulated color deviations over these two steps, there are noticeable changes in the final colors. Why does the controlnet have a color difference? We assumed that the reason is that the pixel space in ControlNet and the latent space in U-Net represent image information in fundamentally different ways. The pixel space deals with the raw pixel values of an image, while the latent space involves a more abstract, high-dimensional representation created by the VAE encoder. For diffusion models, the misalignment between the characteristics and distribution of information in these two spaces can lead to difficulties in maintaining color consistency during the hair transfer process.

Therefore, as shown in Fig.~\ref{fig:method} (right), we improve the ControlNet structure and propose a new variant named \textbf{Latent ControlNet}. Before the image is input into the ControlNet, the image is encoded into the latent space by the VAE encoder and then sent to the trainable copy of U-Net after a new trainable convolutional layer. Finally, we train our \textit{Bald Converter} and \textit{Latent IdentityNet} based on our proposed Latent ControlNet structure and get the best consistency practice.

\begin{figure*}[!t]
    \centering
    \includegraphics[width=0.95\linewidth]{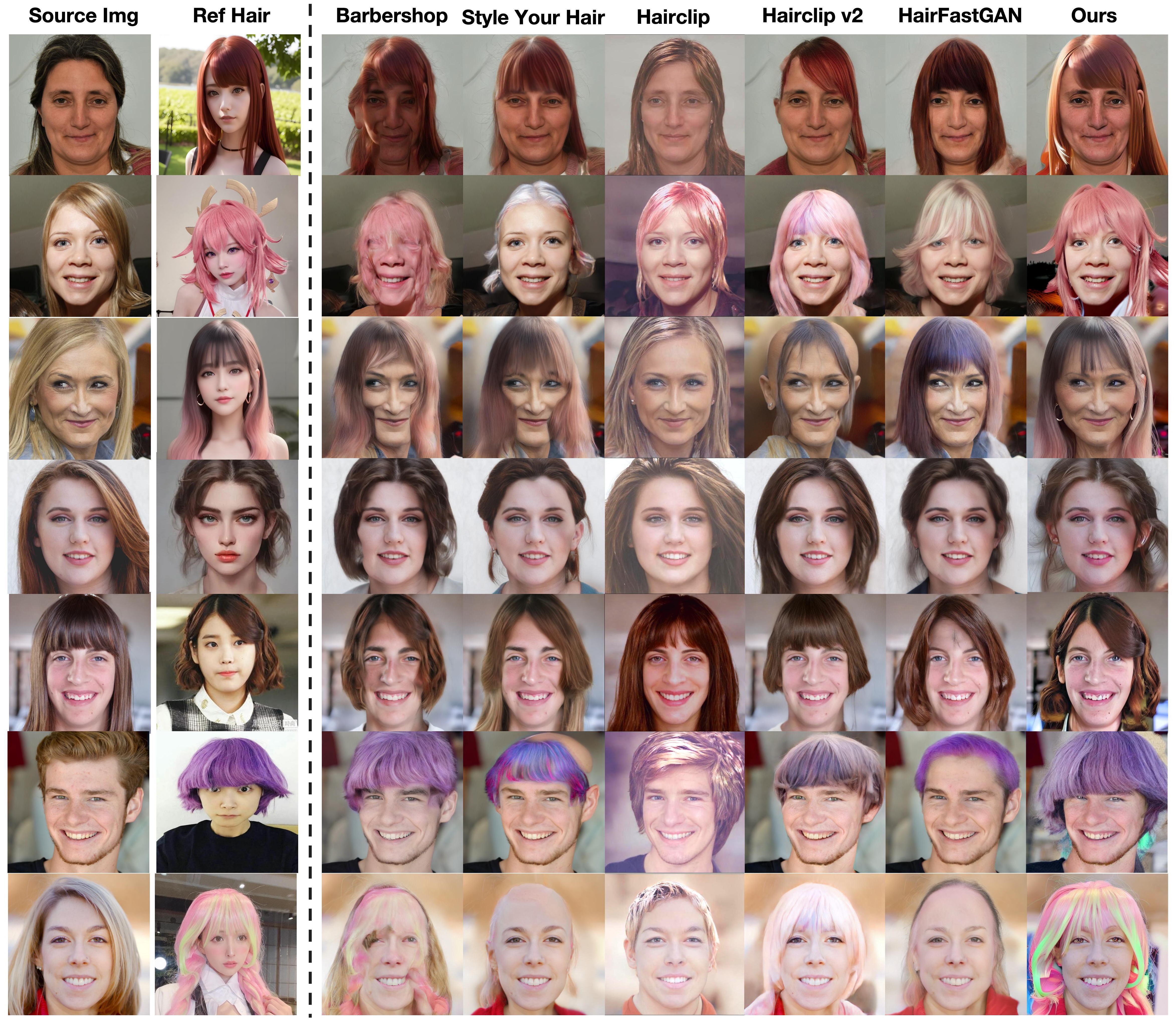}
    \vspace{-0.25cm}
    \caption{Qualitative comparison of different methods. Compared to other approaches, our method achieves more refined and stable hairstyle transfer without the need for precise facial alignment or explicit masks for supervision.}
    \vspace{-0.3cm}
    \label{fig:com}
\end{figure*}

\subsubsection{Hair Extractor.}
Hair transfer needs to transfer the hair in the reference image in a detailed, complete, and accurate way, and our Hair Extraxtor is designed to achieve such transfer. Inspired by recent works~\cite{stablegarment,zhang2024flashface} on reference image-guided generation, we utilize a trainable copy of the U-Net from pre-trained diffusion models as our Hair Extractor. Specifically, we encode our reference image through the Hair Extractor and collect the features of the self-attention layer in each transformer block as detailed hair features. The features are then subsequently injected into the diffusion U-Net through newly added hair cross-attention layers. In each transformer block of the U-Net, we retain the original self-attention layers and add hair cross-attention layers. Detailed hair features are entered into the hair cross-attention layers and serve as the K(key) and V(value) features. Both attention layers in the U-Net share the Q feature. Finally, we simply add the output of hair cross-attention to the output of original self-attention.

\subsection{Model Training}
In the first stage, we train the bald converter using a straightforward approach similar to ControlNet on an existing dataset. This allows us to achieve a highly effective bald converter. In the second stage, we focus on training the Hair Extractor and the Latent IdentityNet. In these two training processes, we have used a variety of augmentations, which are crucial for adapting to real-world scenarios and achieving successful hair transfer. These augmentations include synchronized affine transformations applied to the source image, the bald proxy image, and the target image.

The loss functions for both stages can be mathematically represented as follows:
\begin{equation}
    L(\bm{\theta}) := \mathbb{E}_{\mathbf{x_0}, t, \bm{\epsilon}}\left[\left\|\bm{\epsilon}-\bm{\epsilon_\theta}\left(\mathbf{x_t}, t,\mathbf{c_s},\mathbf{c_r}\right)\right\|_2^2\right],
\end{equation}
where $\mathbf{x_t}$ is a noisy image latent constructed by adding noise $\bm{\epsilon} \in \mathcal{N}(\mathbf{0},\mathbf{1})$  to the image latent $\mathbf{x_0}$ and the network $\bm{\epsilon_\theta(\cdot)}$ is trained to predict the added noise, $\mathbf{c_s},\mathbf{c_r}$ represent the source condition input(source image or bald proxy image), and reference condition input, respectively. When training the bald converter, $\mathbf{c_r}$ is None.

\section{Experiments}
\subsection{Implementation Details} 
We employed Stable Diffusion V1-5 as the pre-trained diffusion model. The model training process utilized a two-stage approach. In the first stage, we trained the Bald Converter using the Non-Hair FFHQ dataset~\cite{Wu_2022_CVPR}. This training was conducted on a single H800 GPU with a batch size of 16 and a learning rate of 5e-5, over a total of 8,000 steps. In the second stage, we trained our Hair Extractor and Latent IdentityNet using the prepared triplet data. This stage was performed on 8 H800 GPUs with a batch size of 8 and a learning rate of 5e-5, over a total of 100,000 steps. During inference, we followed the same two-stage approach as used in training. Both stages utilized the DDIM sampler with 30 sampling steps and the classifier-free guidance scale setting of 1.5.

\setlength{\tabcolsep}{0.9mm}{
\begin{table}[!t]
\centering
\caption{\textbf{Quantitative comparison} of different methods. Metrics that are bold and underlined represent methods that rank 1st and 2nd, respectively.}
\vspace{-0.2cm}
\label{tab1}
\begin{tabular}{*{2}c|ccccc}
\toprule
& Method & CLIP-I $\uparrow$ & FID $\downarrow$ & PSNR $\uparrow$ & SSIM $\uparrow$ & IDS $\uparrow$\\
\midrule
&Barbershop &\underline{0.431} &46.178 &30.397 &0.629 &0.760\\
&HairFastGAN & 0.426&\underline{36.205}  &30.383 &\underline{0.666} &0.762\\
&SYH & 0.429 &41.543 &30.489 &0.645 &0.712\\
&Hairclip &0.391  &44.359 &28.855 &0.615 &0.697\\
&Hairclip v2 &0.419 &37.456 &\textbf{31.552} &0.642&\underline{0.769}\\
\cdashline{1-6}
&Ours &\textbf{0.434} &\textbf{35.128} &\underline{30.980} &\textbf{0.680} &\textbf{0.779} \\
\bottomrule
\end{tabular}
\label{table:1}
\vspace{-0.3cm}
\end{table}
}

\subsection{Evaluation Metrics}
Given a hairstyle reference image, the purpose of hair transfer is to apply the corresponding hairstyle and hair color attributes to the input image. We compare our method with current state-of-the-art methods:
\textit{Barbershop} \cite{zhu2021barbershop}, \textit{SYH} \cite{kim2022styleSYH}, \textit{HairFastGAN} \cite{nikolaev2024hairfastgan}, \textit{Hairclip} \cite{wei2022hairclip}, and \textit{Hairclip v2} \cite{wei2023hairclipv2}. All comparison algorithms use the default parameters from their official implementations.

To provide a thorough and objective assessment of the performance of each algorithm in different aspects of hairstyle transfer, we calculated the \textit{FID} \cite{fid} metrics for the source image and the generated target image. After hairstyle transfer, the identity and background information of the source image and the generated image should be consistent, so we use \textit{SSIM} \cite{wang2004imageSSIM} and \textit{PSNR} to evaluate the identity and background similarity between the source image and the generated target image. Notably, \textit{PSNR} and \textit{SSIM} are calculated at the intersected non-hair regions before and after editing. We also use Insightface \cite{arcface} to evaluate \textit{identity similarity (IDS)} between the original source image and the generated target image. To evaluate hairstyle transfer capabilities, we use CLIP-I\cite{clip} as a metric, which calculates the cosine similarity between the image embeddings of the reference image and the transferred hairstyle image.

\subsection{Experiment Results}
\subsubsection{Qualitative Comparison.}
\begin{figure}[tbp]
    \centering
\includegraphics[width=1\linewidth]{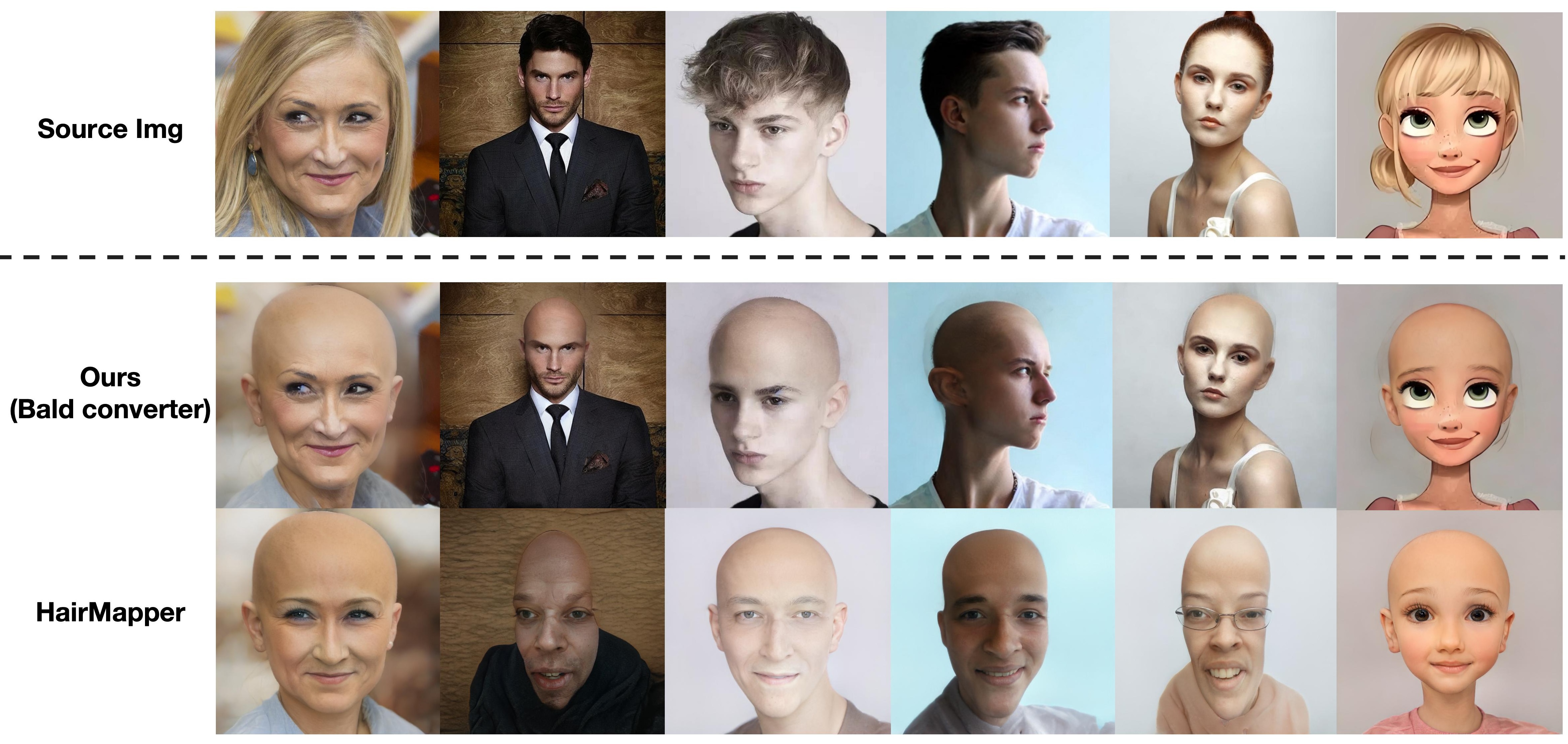}
\vspace{-0.5cm}
    \caption{Visual comparison of hair removal using HairMapper. Our bald converter demonstrates robust performance, effectively converting source images across a wide range of poses, half-body shots, and even animated characters. In contrast, HairMapper struggles with these diverse scenarios, failing to maintain identity and background consistency.}
    \label{fig:bald}
    \vspace{-0.3cm}
\end{figure}

As illustrated in Fig.~\ref{fig:com}, we conducted qualitative comparison experiments on a variety of hairstyles. Overall, our method significantly outperforms other approaches in terms of the refinement and completeness of hairstyle transfer, while also maintaining the background and identity consistency of the source image to a great extent. Among the methods we compared, Barbershop tends to produce chaotic results for complex hairstyle transfers, as observed in the second and fifth rows. Style Your Hair and HairFastGAN perform relatively coarsely in hairstyle transfer, often neglecting the fine details of hair texture and color. Hairclip and Hairclip v2 demonstrate the weakest ability in hairstyle transfer, struggling to accurately transfer the reference hairstyle. In contrast, our method consistently exhibits robust and stable transfer capabilities across different hairstyle styles and colors. Furthermore, We also compared the hair removal capabilities of our Bald Converter with HairMapper in Fig.~\ref{fig:bald}. Our Bald Converter demonstrates robust and reliable performance, effectively transforming source images across various challenging scenarios. In contrast, HairMapper frequently struggles in these contexts, leading to inconsistencies in both identity and background.

\subsubsection{Quantitative Comparison.}

The experiment uses the CelebA-HQ dataset~\cite{karras2017celebra} as experimental data, 2500 face images are randomly selected as input from the CelebA-HQ dataset with an equal number of reference images from the remaining Celeb-HQ dataset. Table~\ref{table:1} shows our quantitative evaluation across different methods. Overall, our method outperforms previous approaches on the majority of metrics, attaining either the top or second position across all evaluated criteria. Specifically, the highest SSIM and IDS scores, along with a notable PSNR ranking, demonstrate that Stable-Hair effectively preserves both the background and identity of the source image during hairstyle transfer. Additionally, the lowest FID score indicates that our method achieves superior visual fidelity and realism in the transferred hairstyles. Achieving the highest CLIP-I score indicates that our method excels at aligning the transferred hairstyle with the reference image, showing strong performance in maintaining the desired style.

\subsubsection{User Study.}
Considering the subjective nature of the hairstyle transfer task, we conducted a comprehensive user study involving 30 volunteers. Specifically, we randomly sampled 20 sets of data from our quantitative experiments and selected 10 popular hairstyles from social media as reference styles, using a corresponding number of source images randomly sampled from FFHQ datasets to create an additional 10 sets of data using various algorithms. This resulted in a total of 30 triplets, each consisting of an original image, a reference image, and the transfer results. As with previous methods \cite{wei2022hairclip}, the test results from different algorithms were randomized. For each test sample, volunteers were asked to choose the best option based on three criteria: transfer accuracy, preservation of unrelated attributes, and visual naturalness. The results shown in Table~\ref{tab2} indicate that our method outperforms the comparison methods in transfer accuracy, preservation of unrelated attributes, and visual naturalness.

\setlength{\tabcolsep}{1mm}{
\begin{table}[!t]
\centering
\caption{\textbf{Quantitative ablation} of different design options. (`pixel' refers to the use of the original ControlNet, while 'latent' denotes the use of the Latent ControlNet.)}
\vspace{-0.3cm}
\label{tab:ab}
\begin{tabular}{*{2}c|ccccc}
\toprule
& Method & CLIP-I $\uparrow$ & FID $\downarrow$ & PSNR $\uparrow$ & SSIM$ \uparrow$ & IDS $\uparrow$\\
\midrule
&Ours (latent) &\textbf{0.434} &\textbf{35.128} &\textbf{30.980} &\textbf{0.680} &\textbf{0.779} \\
&Ours (pixel)&0.423 &39.236 &29.317  &0.668 &0.770 \\
\bottomrule
\end{tabular}
\vspace{-0.3cm}
\end{table}
}

\begin{figure}[!t]
    \centering
    \includegraphics[width=1\linewidth]{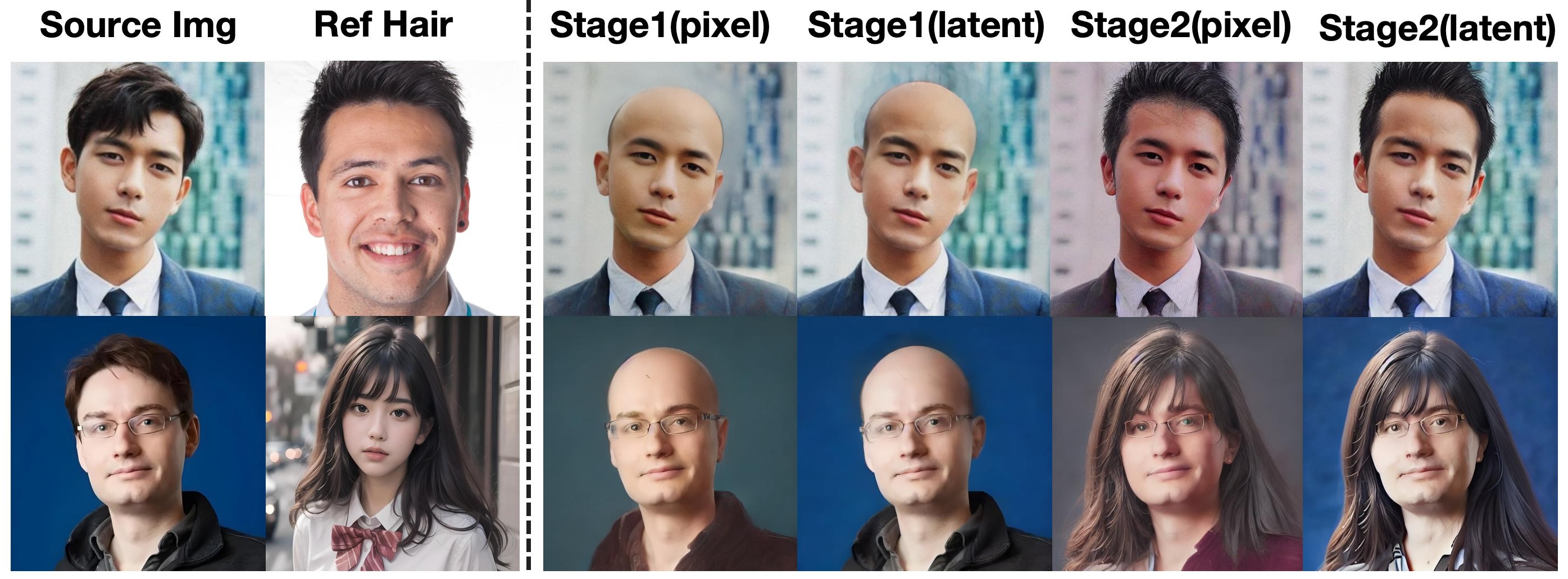}
    \vspace{-0.7cm}
    \caption{Visual ablation study results comparing different design options. The Latent ControlNet clearly maintains color consistency, whereas the original ControlNet introduces color discrepancies at each stage.}
    \vspace{-0.3cm}
    \label{fig:ablation}
\end{figure}

\subsubsection{Ablation Study.}
To thoroughly investigate the role of each module in our method, we conducted systematic ablation. As illustrated in Fig.~\ref{fig:ablation}, it is evident that models trained with pixel-conditioned input using ControlNet often exhibit color discrepancies (as seen in the first and third columns of the results), leading to inconsistencies between the source and target images. Using our proposed Latent ControlNet, which maps the hair removal and transfer process from pixel space to latent space, we effectively eliminate these color inconsistencies, significantly enhancing content preservation. The results of the quantitative ablation are shown in Table~\ref{tab:ab}. Our method consistently outperforms across all metrics, highlighting the superior transfer capabilities and the enhanced preservation of identity and background achieved by the Latent ControlNet.

\begin{figure}[!t]
    \centering
    \includegraphics[width=1\linewidth]{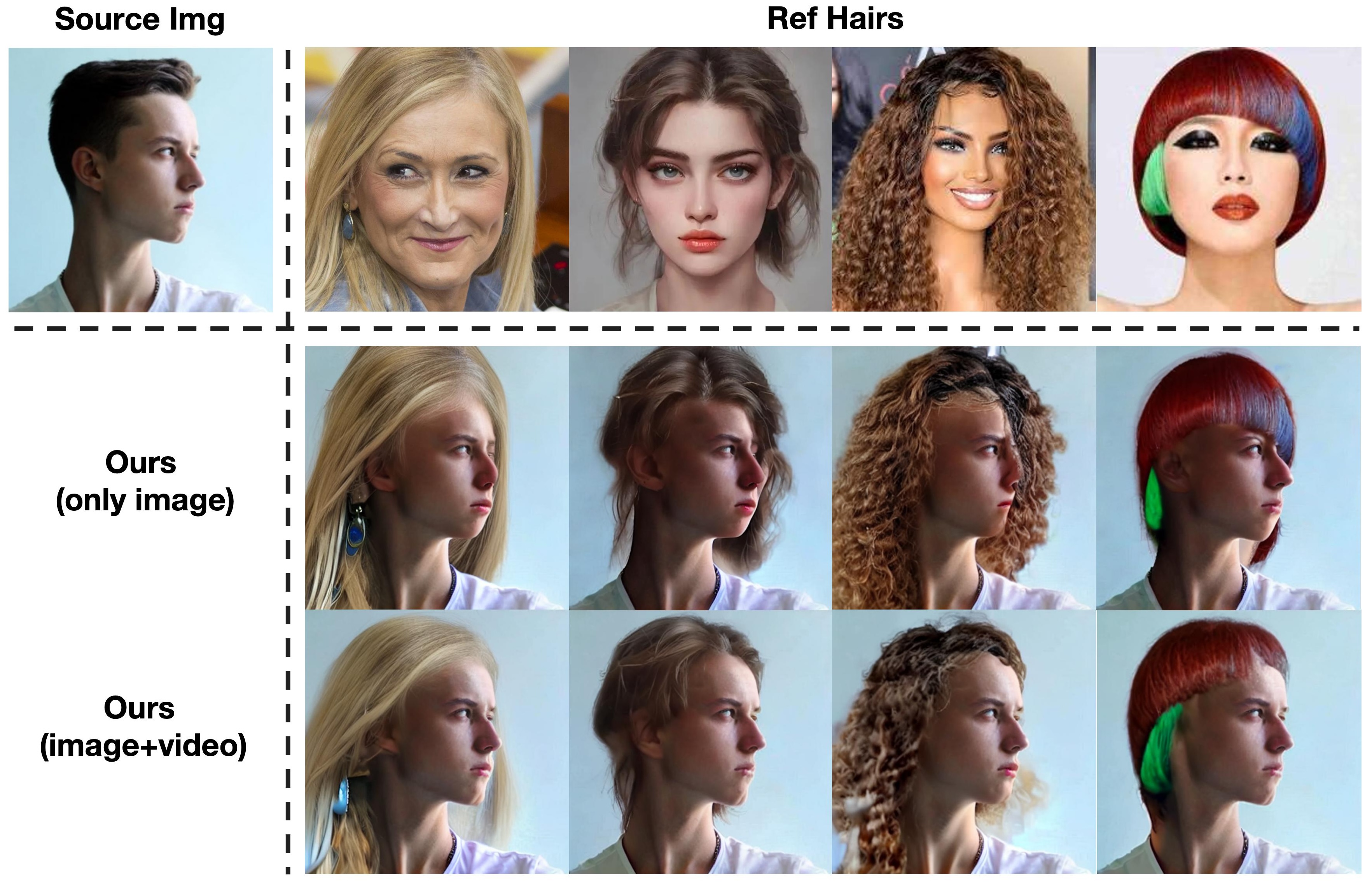}
    \vspace{-0.6cm}
    \caption{\textbf{Visual ablation} results of different training data. When trained exclusively on image data, the model often resorts to a simplistic copy-paste of the hairstyle. However, by integrating video data into the training process, our method gains pose robustness, enabling adaptive and natural hair transfer based on facial orientation.}
    \label{fig:ab_v}
    \vspace{-0.3cm}
\end{figure}

To further investigate the impact of video data on our method, we conducted training using only image data while keeping the same parameter settings. The results of the visual ablation study, as shown in Fig.~\ref{fig:ab_v}, clearly indicate that when trained solely with image data, our method struggles to handle significant variations in facial poses from the source images, leading to a collage-like effect with less natural transitions. In contrast, when trained with a mix of video data, our method demonstrates robustness to changes in facial pose. See the supplements for additional ablation studies.

\setlength{\tabcolsep}{0.6mm}{
\begin{table}[!t]
\centering
\caption{\textbf{User study} on hair transfer. Accuracy denotes the accuracy for hair transfer, Preservation indicates the ability to preserve irrelevant regions and Naturalness denotes the visual realism of the generated image. Our method achieves the best results in all three categories.}
\vspace{-0.3cm}
\begin{tabular}{*{2}c|cccccc}
\toprule
&\multirow{2}*{Metrics}  &Barber  &Hair  &\multirow{2}*{SYH}  &Hair  &Hair &\multirow{2}*{Ours} \\
& &shop &FastGAN & &clip &clipV2 & \\
\midrule
&Accuracy(\%) &18.6  &20.1  &19.2  &5.1  &10.2  &\textbf{26.8}\\
&Preservation(\%) &11.1  &17.3  &20.0  &7.4  &21.8  &\textbf{22.4}\\
&Naturalness(\%) &13.4  &20.2  &15.9  &11.2  &18.7  &\textbf{20.6} \\
\bottomrule
\end{tabular}
\vspace{-0.4cm}
\label{tab2}
\end{table}
}

\section{Conclusions}
In this paper, we introduce Stable-Hair, the first framework to tackle hairstyle transfer using diffusion techniques. This method marks a significant advancement, achieving stable and fine-grained real-world hairstyle transfers previously unattainable. Stable-Hair features a two-stage pipeline. The first stage uses a Bald Converter to transform the source image into a bald proxy image. Following this step, we designed an automated data collection pipeline that leverages the Bald Converter, an inpainting model, and ChatGPT to gather a diverse and large-scale dataset of triplets for training the second stage. In the second stage, a Hair Extractor and a Latent IdentityNet are utilized to accurately and robustly transfer the target hairstyle onto the bald image. Extensive experiments demonstrate that Stable-Hair achieves commercial-grade hairstyle transfer capabilities, setting a new standard in the field.


\clearpage

\begin{center} 
\textbf{Supplementary Materials}
\end{center}

\renewcommand\thesection{\Alph{section}}
\section{Preliminaries}
\label{prel}
\subsubsection{Diffusion Models.}
Diffusion Model (DM)~\cite{ho2020denoising} belongs to the category of generative models that denoise from a Gaussian prior $\mathbf{x_T}$ to the target data distribution $\mathbf{x_0}$ using an iterative denoising procedure. Latent Diffusion Model (LDM)~\cite{rombach2022high} is proposed to model image representations in the autoencoder’s latent space. LDM significantly speeds up the sampling process and facilitates text-to-image generation by incorporating additional text conditions. The LDM loss is as follows:
\begin{equation}
    L_{LDM}(\bm{\theta}) := \mathbb{E}_{\mathbf{x_0}, t, \bm{\epsilon}}\left[\left\|\bm{\epsilon}-\bm{\epsilon_\theta}\left(\mathbf{x_t}, t, \bm{\tau_{\theta}}(\mathbf{c})\right)\right\|_2^2\right],
\end{equation}
where $\mathbf{x_t}$ is an noisy image latent constructed by adding noise $\bm{\epsilon} \in \mathcal{N}(\mathbf{0},\mathbf{1})$  to the image latents $\mathbf{x_0}$ and the network $\bm{\epsilon_\theta(\cdot)}$ is trained to predict the added noise, $\bm{\tau_\theta(\cdot)}$ refers to the BERT text encoder~\cite{bert} used to encodes text description $\mathbf{c_t}$.

Stable Diffusion (SD) is a widely adopted text-to-image diffusion model based on LDM. Compared to LDM, SD is trained on a large LAION~\cite{laion5b} dataset and replaces BERT with the pre-trained CLIP~\cite{clip} text encoder.

\section{Source Data} We utilize these three datasets to collect our training data.

\begin{itemize}
    \item \textbf{FFHQ dataset subset} \cite{karras2019styleFFHQ}: We randomly selected 20,000 images from the FFHQ dataset as the original image data for our training. 
    \item \textbf{CelebV-HQ} \cite{celebvhq}: Large-scale, high-quality video data containing 35,666 clips involving 15,653 identities and 83 manually labeled facial attributes. 
    \item \textbf{NH-FFHQ dataset}: Non-Hair-FFHQ dataset was generated by processing the FFHQ dataset through the hairmapper \cite{Wu_2022_CVPR} method, which is a high-quality image dataset that contains 6,000 non-hair-FFHQ portraits.
\end{itemize}

\setlength{\tabcolsep}{0.3mm}{
\begin{table}[!b]
\centering
\caption{\textbf{Quantitative ablation} of different training data.}
\vspace{-0.3cm}
\label{tab:ab}
\begin{tabular}{*{2}c|ccccc}
\toprule
& Method &CLIP-I$\uparrow$ & FID $\downarrow$ & PSNR $\uparrow$ & SSIM$ \uparrow$ & IDS $\uparrow$\\
\midrule
&Ours (image+video)&\textbf{0.434} &35.128 &\textbf{30.980} &\textbf{0.680} &\textbf{0.779} \\
&Ours (only image) &0.431 &\textbf{33.653} & 30.541 &0.676 &0.771\\
\bottomrule
\end{tabular}
\vspace{-0.3cm}
\end{table}
}

\section{Quantitative Ablation of Training Data}
In this section, we further examine the quantitative impact of incorporating video data into the training set on the performance of hairstyle transfer. As shown in Table \ref{tab:ab}, when the training data consists solely of images, the FID metric shows improvement, while other metrics exhibit slight declines compared to the case where the full dataset is used. This indicates that while there is a slight decline in fidelity and overall image generation quality when video data is incorporated into the training, the model's ability to transfer hairstyles and maintain identity and background consistency is enhanced

\section{Why Use Bald Converter?}

First, removing the original hairstyle eliminates the need for the model to blend the new hairstyle with any remnants of the old one, reducing visual artifacts and resulting in a more natural and visually appealing transfer. Secondly, by standardizing the source images to a bald state, the model can better focus on facial features and structure, enhancing the accuracy and realism of the hairstyle transfer. Additionally, this preprocessing step helps to reduce variability in the training data, improving the model's generalization capability across different hairstyles. Lastly, converting to a bald state simplifies the alignment process, ensuring that the new hairstyle seamlessly integrates with the subject's face, even in cases of diverse facial poses. 

\begin{figure*}[htbp]
    \centering
\includegraphics[width=1\linewidth]{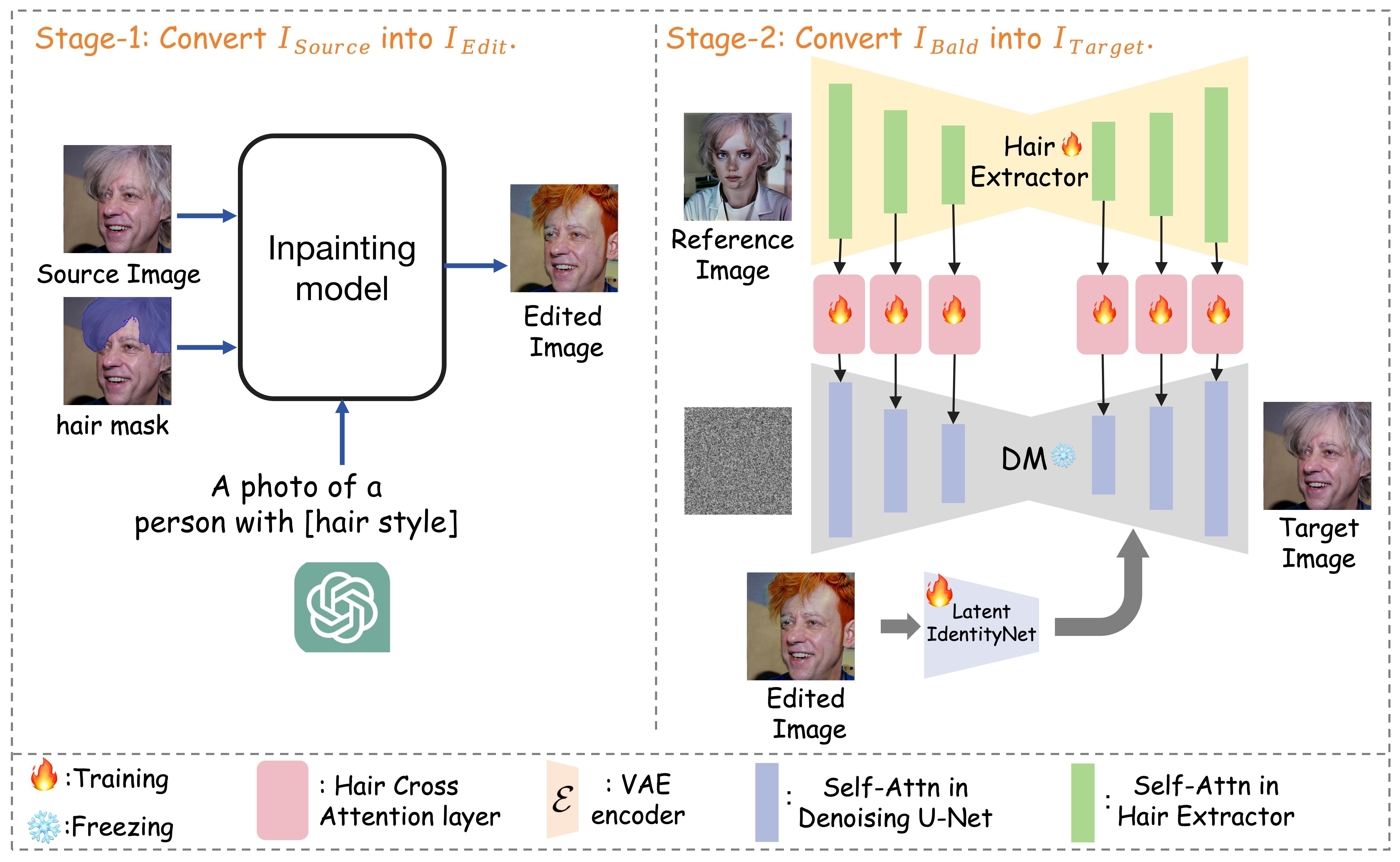}
    \caption{The training framework without the bald converter. (i.e., the source data is not standardized to the bald state)}
    \label{fig:sup_method}
\end{figure*}

\begin{figure*}[htbp]
    \centering
\includegraphics[width=1\linewidth]{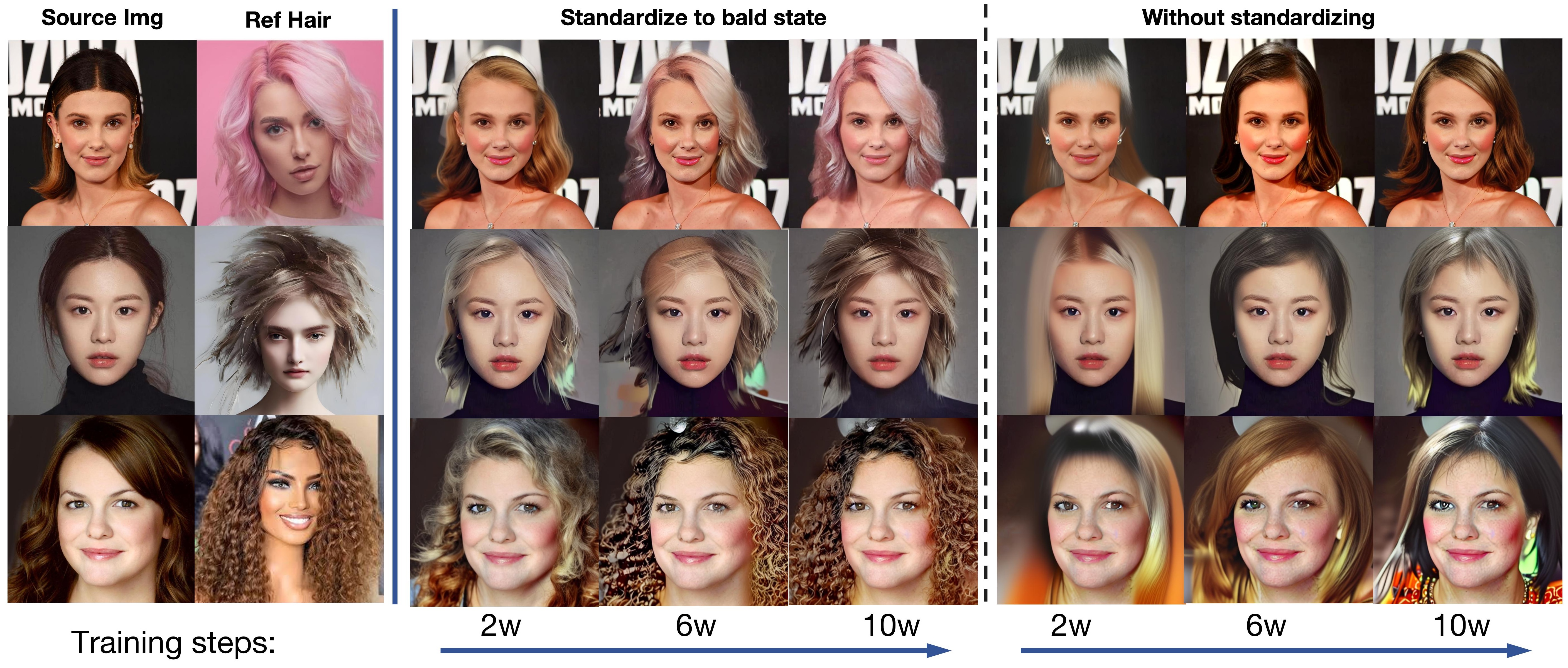}
    \caption{Visual comparison of the results under different data strategies reveals significant differences. When the bald converter is not applied (i.e., the source data is not standardized to the bald state), the model fails to learn how to transfer hairstyles during training and instead generates random hairstyles. However, when the data is standardized to the bald state, the model incrementally learns to transfer high-fidelity and high-precision hairstyles.}
    \label{fig:sup_std}
\end{figure*}

To further demonstrate the importance of standardizing the source image's hairstyle to a bald state in Stage 1, we conducted a set of comparative experiments. In these experiments, rather than standardizing to a bald state as in Stage 1, we randomly edited the source image's hairstyle using an inpainting model and then trained a hair transfer model based on these edited images. The training framework and data references are illustrated in the figure~\ref{fig:sup_method}, with all training settings consistent with those described in our experiments section. The visual comparison results are also shown in the figure~\ref{fig:sup_std}. As evident from the results, our model, which standardizes the hairstyle to a bald state, achieves high-fidelity and precise hair transfer, whereas the model trained on randomly edited hairstyles fails to exhibit any effective hairstyle transfer capability.

\begin{figure*}[htbp]
    \centering
\includegraphics[width=1\linewidth]{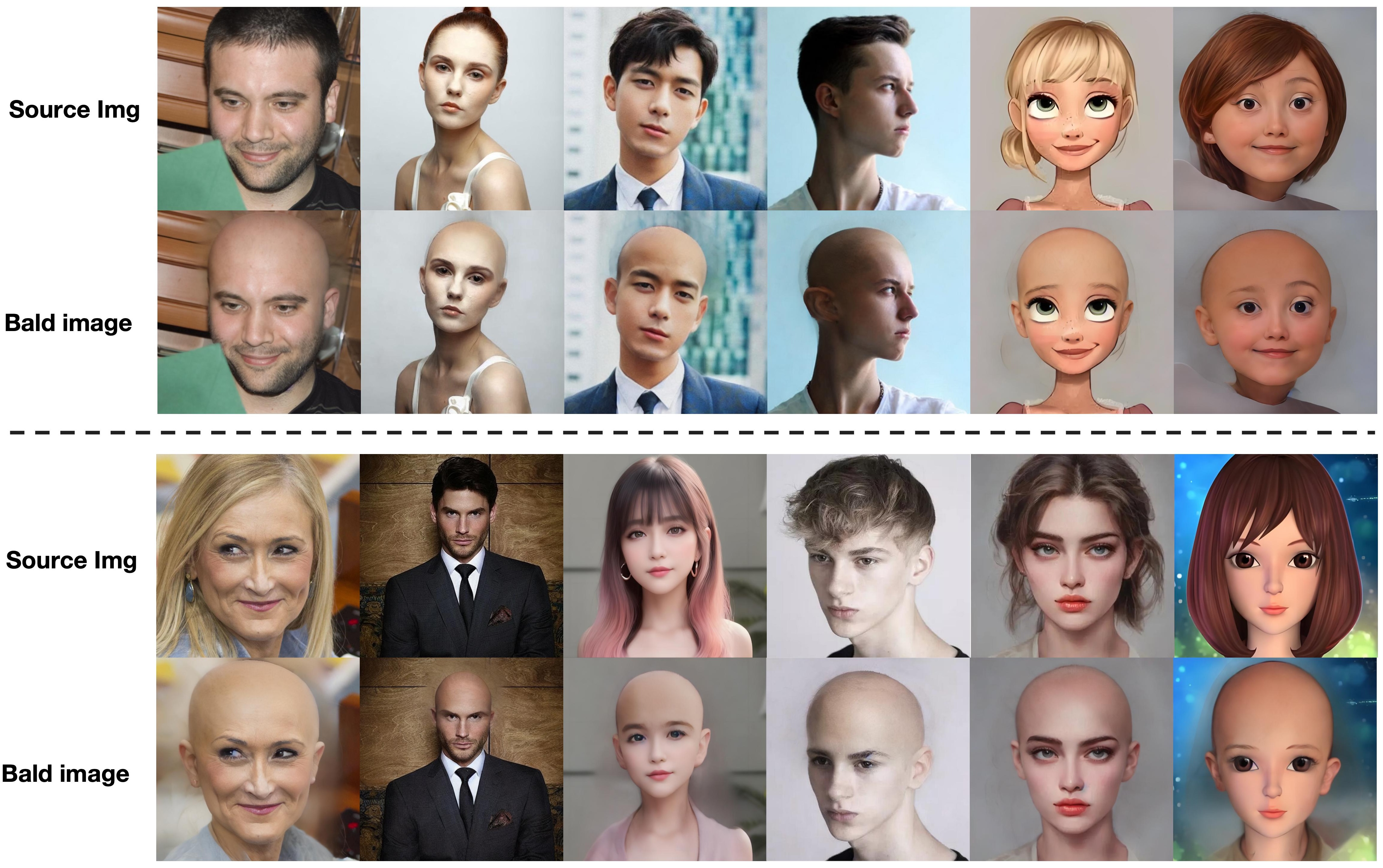}
    \caption{More visual results of hair removal.}
    \label{fig:sup_bald}
\end{figure*}

Additionally, our method achieves hair removal without requiring image alignment or cropping, ensuring that the entire hairstyle is accurately removed and no residual hair remains. This guarantees a cleaner, more consistent starting point for subsequent hairstyle transfer, leading to improved visual fidelity and more realistic results. As shown in Fig. \ref{fig:sup_bald}, our method is capable of removing hair from images with various poses, half-body shots, and even cartoon or animated characters.

\section{More Qualitative Comparison}
\begin{figure*}[htbp]
    \centering
\includegraphics[width=1\linewidth]{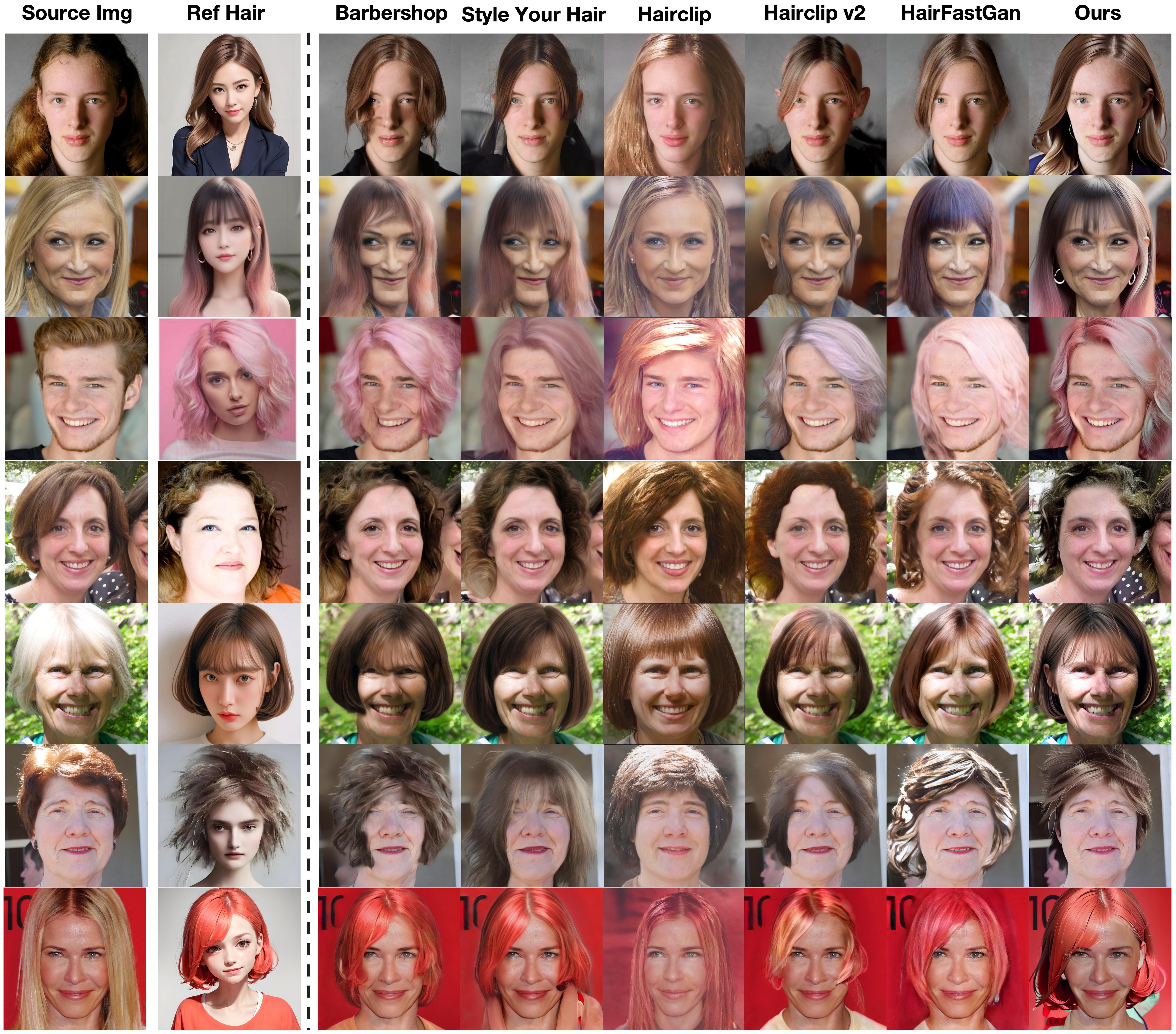}
    \caption{More qualitative comparison results of hair transfer.}
    \label{fig:sup_com}
\end{figure*}

To demonstrate the effectiveness of our method, we present more qualitative comparison results in Fig.~\ref{fig:sup_com}. It is observed our Stable-Hair achieves the most precise and natural transfer. In contrast, all other methods failed to retain details and resulted in unnatural transfers.

\section{More Transfer Results}
\begin{figure*}[htbp]
    \centering
\includegraphics[width=1\linewidth]{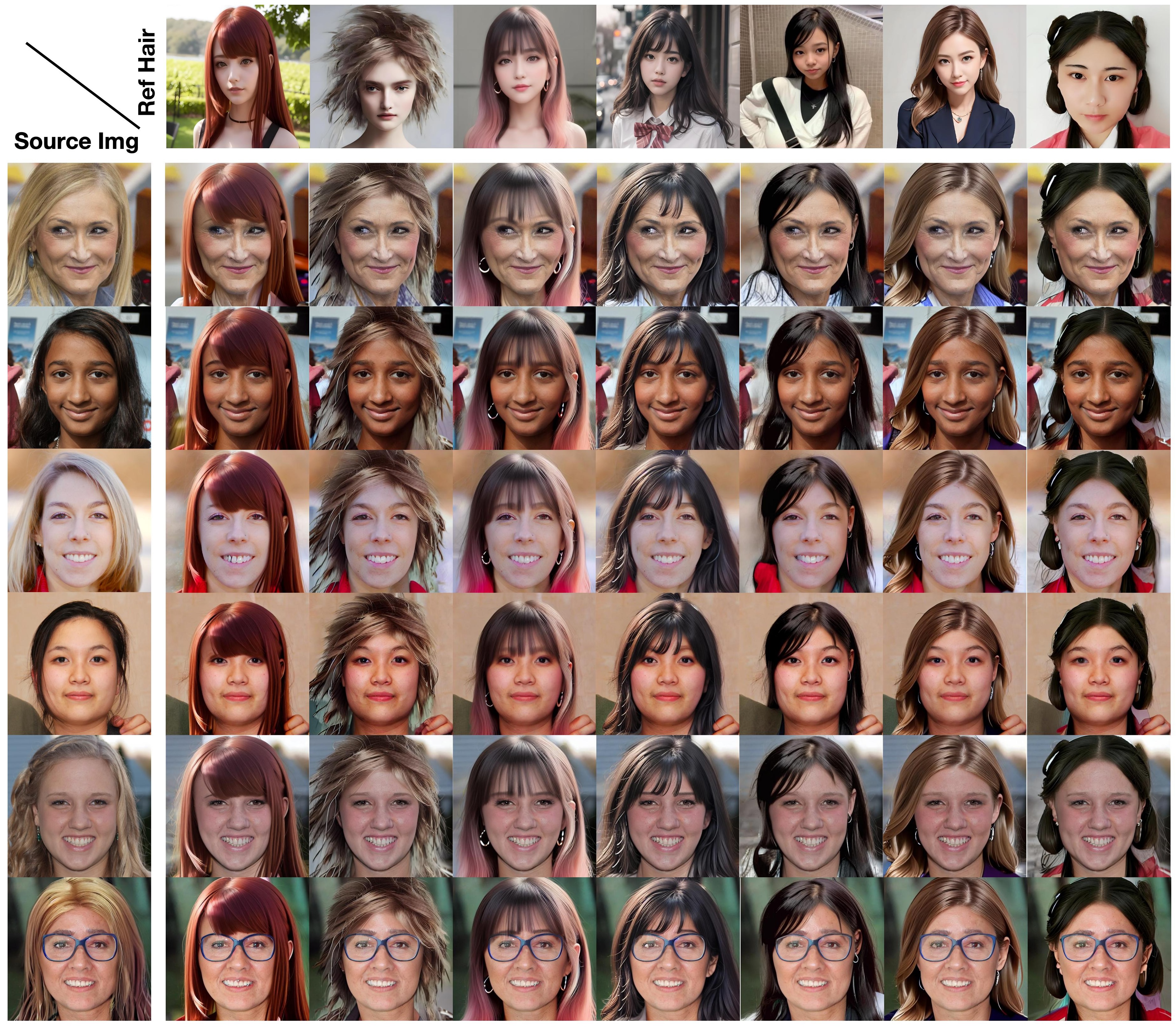}
    \caption{More results of hair transfer for the realistic domain.}
    \label{fig:sup_real}
\end{figure*}

\begin{figure*}[htbp]
    \centering
    \includegraphics[width=1\linewidth]{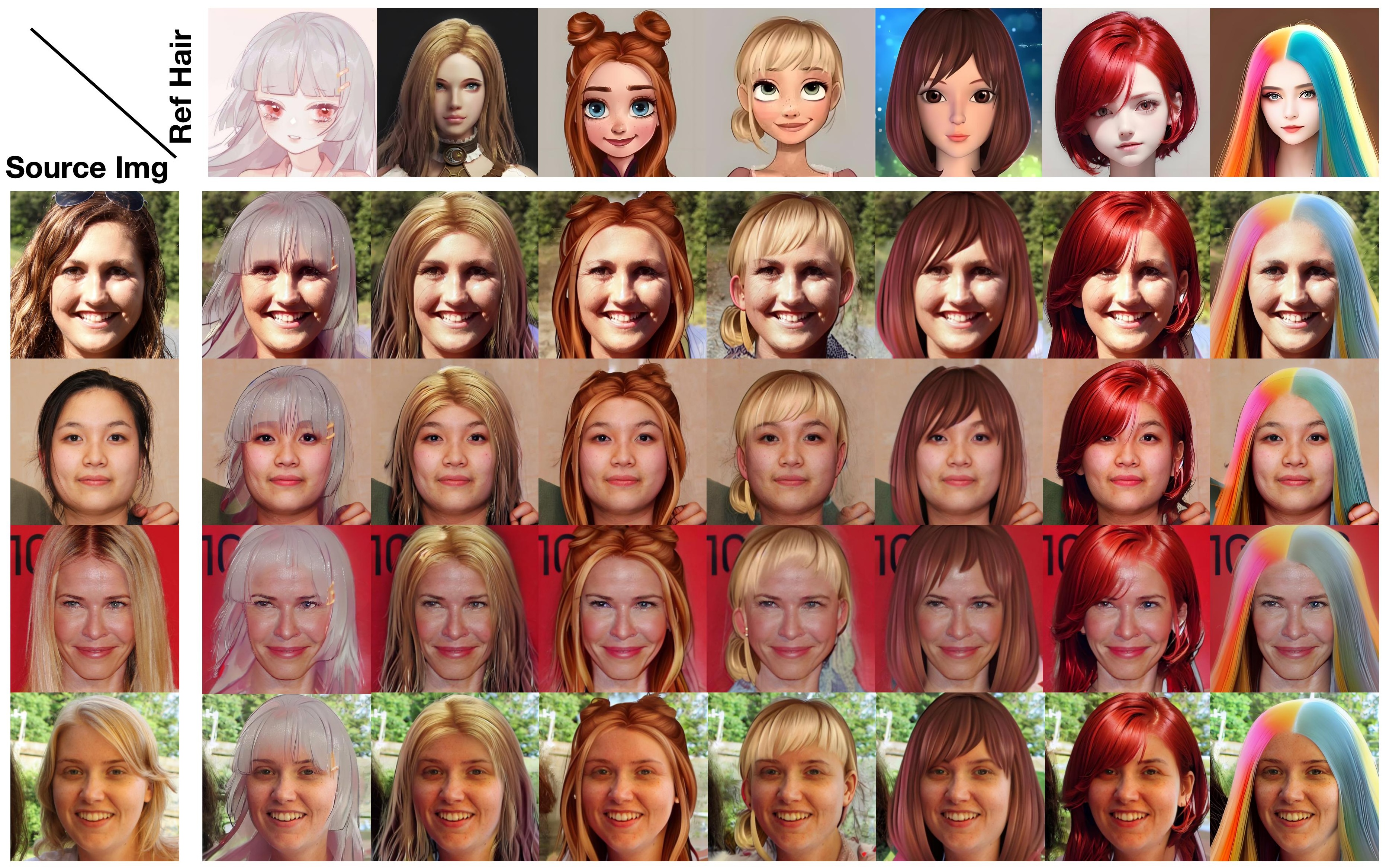}
    \caption{More results of hair transfer for cross-domain.}
    \label{fig:sup_cartoon}
\end{figure*}

In this section, we present additional visual results for hair transfer to demonstrate the robustness and superiority of our approach. As shown in Fig.~\ref{fig:sup_real} and Fig.~\ref{fig:sup_cartoon}, we showcase the effectiveness of our method in hair transfer on realistic domain images and hair transfer on cross-domain images.

\begin{figure}[!t]
    \centering
    \includegraphics[width=1\linewidth]{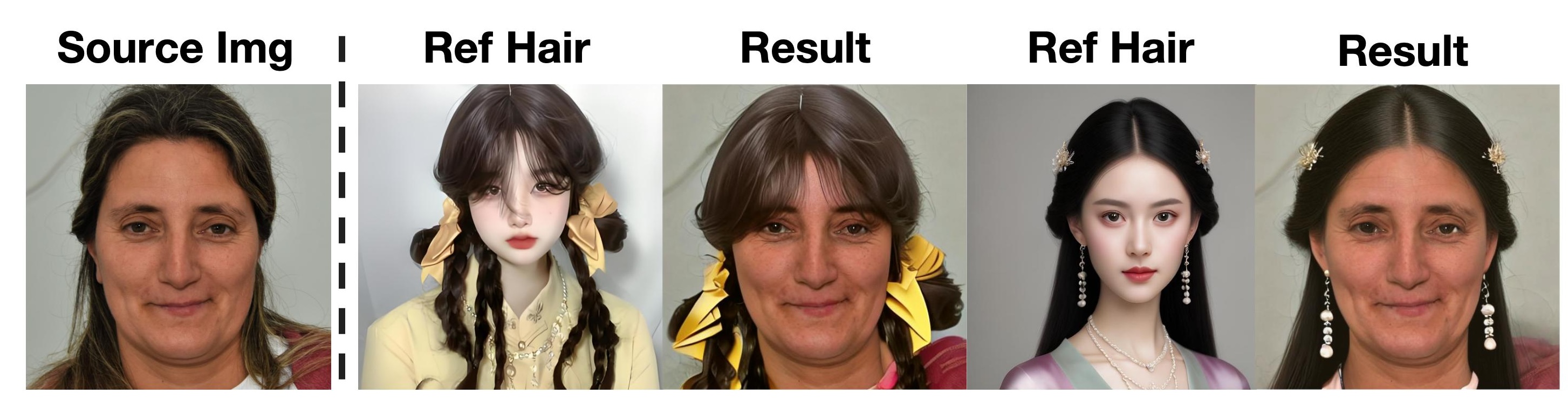}
    \caption{Limitation of our method.}
    \label{fig:sup_lim}
\end{figure}

\section{Limitations}
Due to the limitation of the training data, our method may inadvertently transfer certain hair accessories to the source image (as shown in Fig.~\ref{fig:sup_lim}), which may not be desirable in some scenarios. We plan to address this limitation in future work to achieve more controlled and precise hair transfer.

\section{Broader Impact}
The broader impact of diffusion-based hair style transfer methods is substantial, as they hold the potential to transform the beauty industry by facilitating more efficient and personalized hair styling applications. However, it is crucial to address the ethical considerations associated with this technology, including concerns related to privacy, consent, and the reinforcement of societal beauty standards. We explicitly discourage the unauthorized use of our method to alter the hairstyles in others' photographs. As with any advancing technology, we advocate for a cautious approach in the deployment of diffusion-based hairstyle transfer methods and emphasize the importance of ongoing scrutiny regarding their ethical and legal ramifications.

\bibliographystyle{IEEEtran}

\end{document}